%% file: elsarticle-template-harv.tex
\documentclass[preprint,12pt,authoryear]{elsarticle}

\usepackage{amssymb}
\usepackage{hyperref}
\usepackage{graphicx}
\usepackage{color}
\usepackage{makecell}
\usepackage{pdflscape}
\usepackage{bigstrut}
\usepackage{multirow}
\usepackage{array}
\usepackage{tabularx}
\usepackage{dblfloatfix}
\usepackage{algorithm}
\usepackage[]{mdframed}
\usepackage{algpseudocode}
\usepackage{amsmath}
\usepackage{amssymb}
\usepackage{pifont}
\usepackage{tikz}
\usepackage{subfig}

\usepackage{tikz}
\usepackage{array}
\usepackage{amsmath}
\usetikzlibrary{arrows,positioning,shapes.geometric}
\usepackage[utf8]{inputenc} % allow utf-8 input
\usepackage[T1]{fontenc}    % use 8-bit T1 fonts
\usepackage{hyperref}       % hyperlinks
\usepackage{url}            % simple URL typesetting
\usepackage{booktabs}       % professional-quality tables
\usepackage{amsfonts}       % blackboard math symbols
\usepackage{nicefrac}       % compact symbols for 1/2, etc.
\usepackage{microtype}      % microtypography
\usepackage{lipsum}
\usepackage{fancyhdr}       % header
\usepackage{graphicx}       % graphics
\usepackage{longtable}
\usepackage{caption} 
\graphicspath{{media/}}     % organize your images and other figures under media/ folder
\usepackage{xcolor,colortbl}
\definecolor{Gray}{gray}{0.82}
\definecolor{Gray1}{gray}{0.89}
\journal{Information Systems}

\begin{document}

\begin{frontmatter}



\title{Unfair TOS: An Automated Approach using Customized BERT}


\author{Bathini Sai Akash}

\affiliation{organization={BITS Pilani Hyderabad Campus},%Department and Organization
            addressline={Secunderabad}, 
            city={Hyderabad},
            postcode={500084}, 
            state={Telangana},
            country={India}}

\author{Akshara Kupireddy}

\affiliation{organization={Symbiosis Law School, Hyderabad},%Department and Organization
            addressline={Mamidipalle, Nandigaon}, 
            city={Hyderabad},
            postcode={509217}, 
            state={Telangana},
            country={India}}

\author{Lalita Bhanu Murthy}

\affiliation{organization={BITS Pilani Hyderabad Campus},%Department and Organization
            addressline={Secunderabad}, 
            city={Hyderabad},
            postcode={500084}, 
            state={Telangana},
            country={India}}

\begin{abstract}
\noindent { 
Terms of Service (ToS) form an integral part of any agreement as it defines the legal relationship between a service provider and an end-user. Not only do they establish and delineate reciprocal rights and responsibilities, but they also provide users with information on essential aspects of contracts that pertain to the use of digital spaces. These aspects include a wide range of topics, including limitation of liability, data protection, etc. Users tend to accept the ToS without going through it before using any application or service. Such ignorance puts them in a potentially weaker situation in case any action is required. Existing methodologies for the detection or classification of unfair clauses are however obsolete and show modest performance. 
In this research paper, we present SOTA(State of The Art) results on unfair clause detection from ToS documents based on unprecedented custom BERT Fine-tuning in conjunction with SVC(Support Vector Classifier). The study shows proficient performance with a macro F1-score of 0.922 at unfair clause detection, and superior performance is also shown in the classification of unfair clauses by each tag. Further, a comparative analysis is performed by answering research questions on the Transformer models utilized.  In order to further research and experimentation the code and results are made available on \href{https://github.com/batking24/Unfair-TOS-An-Automated-Approach-based-on-Fine-tuning-BERT-in-conjunction-with-ML}{GitHub}.}
\end{abstract}

% \noindent { 
% Terms of Service (ToS) form an integral part of any agreement as it defines the legal relationship between a service provider and an end-user. Not only do they establish and delineate reciprocal rights and responsibilities, but they also provide users with information on essential aspects of contracts that pertain to the use of digital spaces. These aspects include a wide range of topics, including limitation of liability, data protection, etc. Users tend to accept the ToS without going through it before using any application or service. Such ignorance puts them in a potentially weaker situation in case any action is required. Existing methodologies for the detection or classification of unfair clauses are however obsolete and show modest performance. 
% In this research paper, we present SOTA(State of The Art) results on unfair clause detection from ToS documents based on unprecedented Fine-tuning BERT in integration with SVC(Support Vector Classifier). The study shows proficient performance with a macro F1-score of 0.922 at unfair clause detection, and superior performance is also shown in the classification of unfair clauses by each tag. Further, a comparative analysis is performed by answering research questions on the Transformer models utilized.}


\begin{highlights}
\item ToS defines legal relationships between Service Providers and End-users in digital services.
\item Users often accept ToS without complete comprehension, risking legal implications.
\item Traditional methodologies for ToS unfair clause detection show limited efficacy. Additionally, existing ML methodologies applied to the task are obsolete, failing to perform adequately.
\item Research presents SOTA results in unfair clause detection from ToS documents via integration of custom Fine-tuning BERT with SVC. 

\item Comparative analysis conducted, focusing on Transformer models used, to answer pertinent research questions.
\end{highlights}


\end{frontmatter}



\input{Sections/introduction.tex}
\input{Sections/related_work}

\input{Sections/dataset_desc}
\input{Sections/methodology}

\input{Sections/research_framework}

\input{Sections/results_desc}

\input{Sections/case_study}

\input{Sections/conclusion}
\input{Sections/appendix}

% \section{}
% \label{}








% \end{thebibliography}
% \bibliographystyle{unsrt}  
% \bibliography{references}

\bibliographystyle{ACM-Reference-Format}
\bibliography{references}
\end{document}

%% file: Sections/introduction.tex
\vspace{-15pt}
\section{\MakeUppercase{Introduction}}
\vspace{-5pt}
As consumers, we know that though we are notified on perusing the Terms of Service (hereinafter ToS) before the use of any application or service online, we often accept the conditions without reviewing them thoroughly. Each smartphone typically contains a minimum of 10 to 15 applications, each governed by its respective set of ToS. It is the responsibility of the user to go through the TOS and understand how their personal and sensitive information is dealt with. In simple words, accepting ToS means that we have entered into an agreement with the provider and now we will be governed by these terms.
Various modalities of acceptance are observed in relation to different software and applications: (i) Upon downloading an application onto our devices, we are prompted to consent to the provider's ToS; (ii) While browsing online, upon accessing a web page, a notification appears requesting acceptance of cookies, privacy policy, or ToS; (iii) Some websites implement implicit acceptance, assuming consent is granted through the act of browsing the content of the site, among other approaches. Consumers or end-users do not think twice before accepting these terms because usage of the application at the moment is perceived much more necessary than going through legal implications later. For example, a consumer that has accepted the terms where the company has limited its liability is a complex situation because if the consumer is dissatisfied with the services provided, he can only claim to the extent limited though the damages are much higher. It is also observed that the clauses are quite overwhelming to even erudite users to read and understand as shown in the case of \cite{Dept}. Estimates suggest that merely dedicating time to read ToS or privacy policies could amount to an annual time investment of more than 200 hours per individual as shown in  \cite{cranor}. Considering this, it can be interpreted that a lack of awareness among consumers on their rights and limitations has led to the very concept and core conundrum of this research, “acceptance of unfair terms and conditions”. Further, startups in general don't have a dedicated Legal team for recognition of unfair clauses in the Business-to-Business(B2B) services that are being used. This could lead to detrimental effects in the long run. Automation of such unfair clause detection without any interference from legal experts could have numerous lucrative benefits.

Though countries have adopted laws to prevent unfair trade practices, many companies obligate consumers into accepting their terms which are highly dominant and there is a significant imbalance in the rights and obligations of the parties to the contract. It can also be observed that some jurisdictions do not bar companies from practicing these imbalanced rights over consumers.
Our work, hence, deals with identifying unfair terms in ToS contracts consecutively helping end-users and businesses. We present a Legal-BERT and SVC-based model that identifies clauses based on 8 categories(tags): Limitation of liability, Unilateral termination, Unilateral change, Content removal, Contract by choice/use, Choice of Law, Jurisdiction, and Arbitration. The study works on two tasks. The first task is a classification of each statement into unfair or not based on Unfair Binary as shown in Section(\label{dataset_disc}). This task aims to check for the existence of any clause in a statement but not for a specific unfair tag. The second task aims at the categorization of statements into a specific unfair tag. 
The work has the potential for complete automation with notable superior performance, saving crucial time that can be invested elsewhere for users. Its lucrative impact also reduces the financial burden on companies that employ services to go through these contracts. Most importantly, its high impact would be in the B2C market directly providing keen information and rightly deserved decision-making to fellow customers. 

%% file: Sections/related_work.tex
\vspace{-5pt}
\section{\MakeUppercase{Related Work}}
\vspace{-5pt}
Several ideas in the literature have sought to evaluate online legal documents, such as privacy policies and terms of service, to provide protection for the rights of citizens. Without the inclusion of studies with regard to Privacy Policies such as [\cite{bannihatti5}, \cite{fukushima18}, \cite{harkous20}], little effort has been presented in terms of the TOS. In the following section, we show the literature that aimed at unfair clause detection from TOS documents. 

\cite{lippi2019claudette} revolutionized TOS tagging with unfair clauses with the creation of a dataset with TOS documents of 50 companies all annotated with 8 unfair clause categories. The study also uses ensemble classification methods for the prediction of these classes. Next, the following studies made use of the same dataset for TOS unfair clause detection. A notable study by \cite{guarino2021machine} utilized sentence embeddings from mUSE(\cite{yang2019multilingual}) and further ML classifiers for predictions. The results surpassed the results presented in CLAUDETTE(\cite{lippi2019claudette}).
Lucia et al. proposed Legal-BERT in the study \cite{zheng2021does}. Legal-BERT stems from pre-training vanilla BERT further on 12 GB(53000 data entries) of varied English legal material from many disciplines (such as legislation, court cases, and contracts). 
Next, \cite{mamakas2022processing} proposed adapting a LegalBERT-warm-started Longformer for processing larger texts. This was referred to as LegalLongformer, however, it underperformed as compared to LegalBERT in the case of TOS unfair clause classification. 

Furthermore, the study by \cite{ruggeri2022detecting} investigates configurations for memory-augmented neural networks with a focus on the significance of rationales in context modeling. According to the results, rationales improve classification precision and provide concise, understandable justifications for classifier output that might otherwise be ambiguous. The study further added 50 TOS documents to the dataset annotated by \cite{lippi2019claudette} leading to a total of 100 TOS documents and focused on only 5 categories of unfair clauses. 
As an extension to this, a recent study by \cite{ruggeri2022detecting}  suggests an addition to transformer models, using external memory to store and then apply natural language interpretations to explain classification outcomes. 

Finally, \cite{xu2022attack} worked on analysis using all-encompassing adversarial triggers to attack an unfair-clause detector. Studies reveal that any minor modification to the text can significantly lower the detection performance.  Our study works on achieving SOTA(State Of The Art) performance at TOS classification of unfair clauses on the dataset presented by \cite{lippi2019claudette}.

%% file: Sections/dataset_desc.tex
\vspace{-6pt}
\section{\MakeUppercase{Background and Dataset}} 
\vspace{-2pt}
AI and technology have recently shaped the legal field by making it easier to understand clauses, rules, and the law itself. This is also evident in the recent hike in the employment of lawyers with tech backgrounds who are well-versed in tech law. These lawyers increase the efficiency of working of law because, through the perspective of science, interpretation, comprehension, and explanation are the three most crucial structures for the proper functioning of a model(\cite{loosforbg1}). Our work deals with these three components by comprehending the law in simple terms and then interpreting and giving accurate outcomes with explanations. 
Companies exploit their customers by causing substantial injury, by not giving the option to negotiate ToS, and by countervailing the unfair clauses with benefits or additional services provided to the customers. The legal system also cannot provide complete assistance with these issues as it is limited to the view that both the parties, the companies, and the end users, should be dealt with in fairness adhering to audi alteram partem. This standpoint can also be observed in the case of \cite{caselaw1}, the Upper House of the Indian Parliament was of the opinion that “the principles against the exclusion of liability clauses are not applicable in their full vigor when considering the facet of mere limitation of liability clauses.” A similar judgment was held in \cite{caselaw2}. Further, in \cite{caselaw3} the courts held that though there is a dominant arbitration clause in favor of one of the parties, it cannot be rendered invalid. But in September 2019, one of the giants- Amazon, was fined 4 million Euros for including unfair clauses such as unilateral termination, unilateral change, limitation of liability, etc. It was also ordered that its “standard terms of service” be amended within 180 days.  
Now that we observe, these clauses are determined as fair and unfair case by case on approaching the courts, and there is no standard law that protects the consumers. In order to increase consumers’ awareness and maintain contractual security, our research suggests establishing a precise description of unfairness in the Terms of Service. This description will be determined by a comprehensive measure of unfairness, which includes eight different categories, which later evaluates the number of unfair provisions incorporated in a company’s ToS. Additionally, the measure will underscore these terms depending on the brunt of the actual concerns of consumers.
\vspace{-1pt}
\subsection{\textbf{Dataset}} \label{dataset_disc}
% \vspace{-5pt}
Our work makes use of the dataset from \cite{lippi2019claudette}. It consists of 8+1 classes. Eight classes are unfair clause classes of each category, and the last class is annotated positive if one or more clauses of any category are present in the statement. This class is referred to as Unfair Binary. An explanation of each unfair category is given below:

% \begin{enumerate}
%     \renewcommand{\labelenumi}{(\theenumi)}
%     \item \textbf{Limitation of Liability}[ltd]: This category in ToS is where companies either completely exclude liability or cap the amount/loss which they are actually liable for. Any clause that explicitly mentioned that the company was liable for certain acts was identified as fair \textbf{(0)}. Instances where the companies capped or excluded their liability were marked as unfair \textbf{(1)}. Example clause: “7(c)neither we nor any of our affiliates or licensors will be liable for any indirect, incidental, special, consequential or exemplary damages, including damages for loss of profits, goodwill, use, or data or other losses, even if we have been advised of the possibility of such damages.” OpenAi. 
    
%     \item \textbf{Unilateral termination}[ter]: This category in ToS is when the service provider retains the right to terminate/suspend a part/clause of ToS or the contract as a whole, temporarily or permanently. Such a termination/suspension could be for any/no reason whatsoever. All the clauses that gave service providers the right were marked as unfair \textbf{(1)}. Example clause: “Any breach of the following restrictions constitutes a breach of the Terms of Use and entitles Ferrari to unilateral termination of the Agreement.” Ferrari. 
%     \item Item 3
% \end{enumerate}

\begin{enumerate}
    \renewcommand{\labelenumi}{(\theenumi)}
    \item \textbf{Limitation of Liability}[ltd]: This category in ToS is where companies either completely exclude liability or cap the amount/loss which they are actually liable for. Any clause that explicitly mentioned that the company was liable for certain acts was identified as fair \textbf{(0)}. Instances where the companies capped or excluded their liability were marked as unfair \textbf{(1)}. Example clause: “7(c)neither we nor any of our affiliates or licensors will be liable for any indirect, incidental, special, consequential or exemplary damages, including damages for loss of profits, goodwill, use, or data or other losses, even if we have been advised of the possibility of such damages.” OpenAi. 
    
    \item \textbf{Unilateral termination}[ter]: This category in ToS is when the service provider retains the right to terminate/suspend a part/clause of ToS or the contract as a whole, temporarily or permanently. Such a termination/suspension could be for any/no reason whatsoever. All the clauses that gave service providers the right were marked as unfair \textbf{(1)}. Example clause: “Any breach of the following restrictions constitutes a breach of the Terms of Use and entitles Ferrari to unilateral termination of the Agreement.” Ferrari.

    \item \textbf{Unilateral change}[ch]: This category in ToS gives the service provider the right to amend or change the terms of the contract or the contract as a whole. Such a change could be for any/no reason whatsoever. All the clauses that gave service provider such right were marked as unfair \textbf{(1)}. Example clause: “You should look at these Terms regularly, which are posted on the Hulu Site at hulu.com/terms. If we make a material change to these Terms, we will notify you by posting a notice on the Hulu Site.” Hulu.

    \item \textbf{Content removal}[cr]: This category in ToS gives the service provider the right to alter/remove any content provided at a point of time with or without any notice. Such a removal could be for any/no reason whatsoever. All the clauses that gave service providers such rights were marked as unfair \textbf{(1)}. Example clause: “You agree that from time to time we may remove the service for indefinite periods of time or cancel the service at any time, without notice to you.” TLC.

    \item \textbf{Contract by using}[use]: This category of ToS stipulates how the customer is legally obligated to adhere to the terms and conditions of utilizing a certain service that is provided by the service provider, regardless of necessitating explicit acknowledgment or acceptance to the conditions of use. All the clauses that gave service providers such rights were marked as unfair \textbf{(1)}. Example clause: “The Terms include our Service Terms, Sharing \& Publication Policy, Usage Policies, and other documentation, guidelines, or policies we may provide in writing. By using our Services, you agree to these Terms.” OpenAi.

    \item \textbf{Choice of law}[law]: This category of ToS lets the service provider specify which state’s law shall govern the contract and which shall prevail if there is any dispute arising out of the contract. All the clauses mentioning the choice of law is that of the consumer’s country was marked as fair \textbf{(0)}. All the clauses that prefer or specify otherwise were marked as unfair \textbf{(1)}. Example clause: “These Terms of Service and any separate agreements whereby we provide you Services shall be governed by and construed in accordance with the laws of South Africa.” TLC.

    \item \textbf{Jurisdiction}[j]: This category of ToS specifies under which jurisdiction the disputes shall be addressed regardless of where the cause of action arose. All the clauses mentioning the jurisdiction as that of the consumer’s residence/country were marked as fair \textbf{(0)}. All the clauses that specify otherwise were marked as unfair \textbf{(1)}. Example clause: “To the extent that the arbitration provision outlined in Section 13 is not applicable, you and Hulu agree to submit to the exclusive jurisdiction of the courts located in the Los Angeles County of the State of California.” Hulu.

    \item \textbf{Arbitration}[a]: This category of ToS mentions arbitration as the first step before approaching the courts or makes arbitration the sole relief. All the clauses mentioning the arbitration as a mere choice were marked as fair \textbf{(0)}. All the clauses that specify otherwise in terms of place of residence or service provider discretion were marked as unfair \textbf{(1)}. Example clause: “In the unlikely event that an issue between us remains unresolved, please note that THESE TERMS REQUIRE ARBITRATION ON AN INDIVIDUAL BASIS, RATHER THAN JURY TRIALS OR CLASS ACTIONS.” Hulu.

    \item \textbf{Unfair binary}[ub]: This category of ToS marks sentences with at least one above unfair clauses as \textbf{1} and all fair/neutral sentences as \textbf{0}. 
\end{enumerate}

Example clauses for each category and statistics of the dataset are presented in Table(\ref{unfair_spread}). In total, there were 9414 statements each annotated into eight categories.

\begin{flushleft}
\begin{table*}[!h]
\caption{ Spread for each unfair category in the Dataset}\label{unfair_spread}
\centering
\scalebox{0.92}{
\begin{tabular}{|c|c|c|}\hline

\hline
\bf Unfair Tag & \bf Clause/Category  &\bf Spread \\\hline
ltd & Limitation of Liability & 296 \\
ter & Unilateral termination & 236 \\
ch & Unilateral change & 188 \\
cr & Content removal & 118 \\
use & Contract by using & 117 \\
law & Choice of law & 70 \\
j & Jurisdiction & 68 \\
a & Arbitration & 44 \\
ub & Unfair binary & 1137 \\\hline
% Loos Et al.\cite{loos2016wanted} & SVM+mUSE & & 0.86 & 0.86 & 0.87 

\end{tabular}}
\end{table*}
\end{flushleft}

% \section{Dataset} \label{Dataset_creation and processing}
% In machine learning, intelligence is extracted from the right data. Prediction performance is remarkably related to data quality. Data shortages are a very evident problem in numerous fields of machine learning.The current state-of-the-art research was developed on scarce datasets due to labeled data constraints. On inspection, the research identified three standard labeled datasets for predicting personality. \\
%  We utilized the Essays dataset \cite{pennebaker1999linguistic}, which includes 2468 student essays annotated with OCEAN Big Five personality and Kaggle MTBI dataset\cite{jolly2017myers}  with  8675 data points gathered from the PersonalityCafe forum. Finally, Our study's myPersonality dataset included a sample of personality ratings using Facebook profile information. The information was gathered by Schwartz et al. \cite{tadesse2018personality} 

%% file: Sections/methodology.tex
\vspace{-8pt}
\section{\MakeUppercase{Fine tuning Pre-Trained Language Models}}
\vspace{-3pt}
The development of "pretrained" (or self-supervised) language models, beginning with Google's BERT model, is considered one of the most important developments in natural language processing (NLP). The incorporation of these exceptional models in the field of law has however been belated. In this study, we first explore multiple transformer architectures in the field of Law for unfair clause detection. 

\textbf{\textit{Legal-BERT}}.
The Bidirectional Encoder Representations from Transformers (BERT, for short) is a comprehensive deep learning-based language model. A transformer model with 12 stacked encoders is at the foundation of the BERT architectural framework. As used in previous studies, unlike context-independent methods like words2vec, Tf-Idf, GloVe, and Elmo, which create single-word embeddings without taking into account global context, BERT employs embeddings that dynamically adjust depending on the contextual use of a particular word across distinct phrases.

% There are two versions of Vannila BERT: In contrast to $ BERT LARGE $, which has 24 encoders and 16 self-attention heads, $ BERT BASE $ has 12 encoders and 12 bi-directional self-attention heads.

The research made use of a further pre-trained BERT base model called Legal Bert. Legal-BERT is a set of BERT models for the legal domain that is meant to support applications in computational law, legal technology, and legal NLP research(\cite{zheng2021does}) Pre-trained on gathered 12 GB(53000 data entries) of varied English legal material(such as legislation, court cases, and contracts), it shows more proficient results in the legal fields as compared to vanilla bert.  

\textbf{\textit{ELECTRA}}
After BERT, a revolutionary pretraining method ELECTRA trains the generator and discriminator transformer models. The generator is trained as a masked language model since its function is to substitute tokens in a sequence. The discriminator attempts to determine which tokens in the chain were substituted by the generator. ELECTRA generically outperforms BERT requiring lesser training data(\cite{clark2020electra}). The study made use of \textit{google/electra-base-discriminator}. 

\textbf{\textit{DeBERTa}} 
DeBERTa is a brand-new model that improves RoBERT and BERT by adding disentangled attention and a better mask decoder for pretraining. Disentangled attention, which represents each word's content and location using two vectors, and an improved mask decoder, which takes the place of the softmax layer and predicts masked tokens, are the two main strategies used(\cite{he2020deberta}). The study made use of \textit{microsoft/deberta-base}. DeBERTa being a lesser-known model, its performance in the field of Law is yet to be tested and analyzed.

\textbf{\textit{DistilBERT}}
DistilBERT is a faster and more compact version of the transformers model proposed by \cite{sanh2019distilbert}. It was pre-trained using the same corpus while being guided by the BERT basic model, which served as a teacher. BERT's performance was not intended to be maximized via DistilBERT. Instead, it seeks to maximize training optimization while minimizing the enormous size of BERT (and having 110M and 340M parameters, respectively). 

\textbf{\textit{DistilRoBERTa }}
The refined version of the RoBERTa base is called \href{https://huggingface.co/distilroberta-base }{DistilRoBERTa}. DistilBERT and DistilRoBERTa have similar training. Since RoBERTa used alternative hyperparameters during training, it performs notably better than BERT.  The model has six layers, seven hundred sixty-eight dimensions, twelve heads, and 82 million parameters (compared to 125M parameters for RoBERTa-base). 
% \vspace{-5pt}
\subsection{\textbf{Custom Fine tune}} \label{custom_finetune}
% \vspace{-2pt}
The study first aims to understand the raw performance of pre-trined language models. To do this, we create custom architectures that incorporate Dense Perceptron layers over the transformer stack to get binary output of unfair clause classification. After the architecture is built, fine-tuning is performed for each model, respectively, and weights are tuned for the task. The performance and architectures are shown in table\ref{fine_tune_arch}. We test with five epochs, batch size in [8,16,32], and learning rate [1e-5,2e-5,4e-5]. In the custom architecture shown, Dense(x) is a perceptron layer where x is the input neuron size. If it is the last Dense layer then output neurons would be 2(Binary Classification). Next, dr(x) is a dropout layer with p=x. 
When employing pre-trained language representation models like BERT, the text does not need dedicated \textit{preprocessing}, and the removal of tokens shows detrimental effects elucidated by \cite{zhong2023revisiting}. In specific, it employs a multi-head self-attention system to utilize all of the information in a phrase, including punctuation and stop words, from a variety of angles. Further, the dataset under consideration is already pre-processed with the removal of unnecessary characters.

\begin{table*}[h!]
\centering
\caption{Performance of Custom Transformer Architectures}\label{fine_tune_arch}
% \centering
% \renewcommand{\arraystretch}{1.8}
\scalebox{0.8}{
\begin{tabular}{|lccccc|}\hline
\hline
\bf  Model &\bf Hyper-parameters & \bf Custom Arch & \bf P & \bf R & \bf F1  \\\hline
Legal-BERT & bs=8,lr=1e-5  & dr(0.1)+Dense(768) & 0.835 & 0.88 & 0.858 \\
Electra & bs=8,lr=2e-5 & Dense(768)+dr(0.1)+Dense(768) & 0.879 &0.848 & 0.863 \\
DeBERTa & bs=16,lr=1e-5 & dr(0.1)+Dense(768) & 0.867 &0.840 & 0.857 \\
DistilBERT & bs=16,lr=4e-5 & Dense(768)+Dense(768)+dr(0.2) & 0.838 &0.816 & 0.826 \\
DistilRoBERT & bs=16,lr=2e-5 & Dense(768)+dr(0.1)+Dense(768) & 0.842 &0.825 & 0.833 \\
\hline
\end{tabular}}
\end{table*}

On inspection of table\ref{fine_tune_arch}, there's a notable difference in the performance of Distil models over traditional transformer models. Further, the performance of ELECTRA slightly improves over Legal-BERT although Legal-BERT was pre-trained(NSP) in the Legal domain. The reason for this could be because of the low domain specificity of TOS in legal documents trained on BERT as mentioned in \cite{zheng2021does}. Hence, we consider Legal-BERT, Electra, and DeBERTa for further testing.

%% file: Sections/research_framework.tex
\vspace{-8pt}
\section{ \MakeUppercase{Research Framework}}  \label{resframework}
\vspace{-2pt}
The following section shows the proposed design and techniques used for  Unfair Binary prediction(Task-1).
\vspace{-2pt}
\subsection{\textbf{Research Methodology}}
% \vspace{-5pt}
While working with transformer models the following prospective methodologies were used in the research are as follows:

\begin{enumerate}
  \item Generation of Vannila LM embeddings\textit{(\textbf{Baseline})}
  \item Further pre-training of LM on Target Corpus\textit{(\textbf{Pretarined})}: The transformer models are pre-trained on TOS dataset. Even though the dataset is considerably smaller compared to the extensive pre-training that was already performed, studies show that marginal performance gains could be achieved\cite{araci2019finbert}. For transformer models, two forms of pre-training exist, Masked Language Modeling (MLM) pre-training and Next Sentence Prediction(NSP). The pre-training performed for Legal-BERT and DeBERTa was MLM pre-training without NSP pre-training. For Electra, (MLM+NSP) pre-training over \textit{electra-base-discriminator} and MLM per-training over \textit{electra-base-generator} were performed separately. But the latter performed better so we utilized it for the research.
  \item Fine-tuning Custom models and saving base model\textit{(\textbf{Fine-tune-base})}: The approach makes use of custom models shown in Section(\ref{custom_finetune}) to fine-tune LM's on train data set(0.15 split) and save only fine-tuned base model without perceptron layers(Figure(\ref{fine_tune_arch}). This unprecedented approach could show prospective results considering the weight optimization due to the fine-tuning of LMs to the task.
  
  \item Pre-training LM's as in approach 2 and then fine-tuning as in approach 3\textit{(\textbf{Pretrained-finetuned})}

\end{enumerate}

\vspace{-2pt}
\subsection{\textbf{Embedding Generation}}
% \vspace{-5pt}
While sentence embeddings are generated from word embeddings the study inspects 8 different embedding generation modes and the best-performing mode is later comprehended. Electra, Legal-BERT, and DeBERTa have 12 transformer layers and each word embedding is a 768-valued vector. The first value in the word vector set for any sentence is the  'cls' token. The 8 embedding modes are shown in Table(\ref{table_embed_modes}). The original authors of BERT, \cite{devlin2018bert} perform ablation studies on combinations of word embeddings which instigated our research. Further, To account for the different modes and variability, probing tests have been conducted to analyse BERT layer semantic similarity, NER, etc(\cite{wallat2023probing}).

\begin{table}[h!]
\centering
\caption{ Different modes for sentence embedding calculation}\label{table_embed_modes}
% \centering
% \renewcommand{\arraystretch}{1.8}
\scalebox{0.93}{
\begin{tabular}{|cl|}\hline
\hline
\bf  Mode Number  \ \ &\bf Explanation  \\\hline
Mode-1 &  Mean only 12th layer \\
Mode-2  & Mean all 13 layers(12\textbf{+}embed)   \\
Mode-3  & Mean 9 to 12 layers   \\
Mode-4  & Mean 7 to 12 layers \\
Mode-5 &  Cls only 12th layer \\
Mode-6  & Cls all 13 layers(12\textbf{+}embed)   \\
Mode-7  & Cls 9 to 12 layers   \\
Mode-8  & Cls 7 to 12 layers \\
\hline
\end{tabular}}
\end{table}
\begin{figure}[!h]
\centering
\includegraphics[scale=0.43]{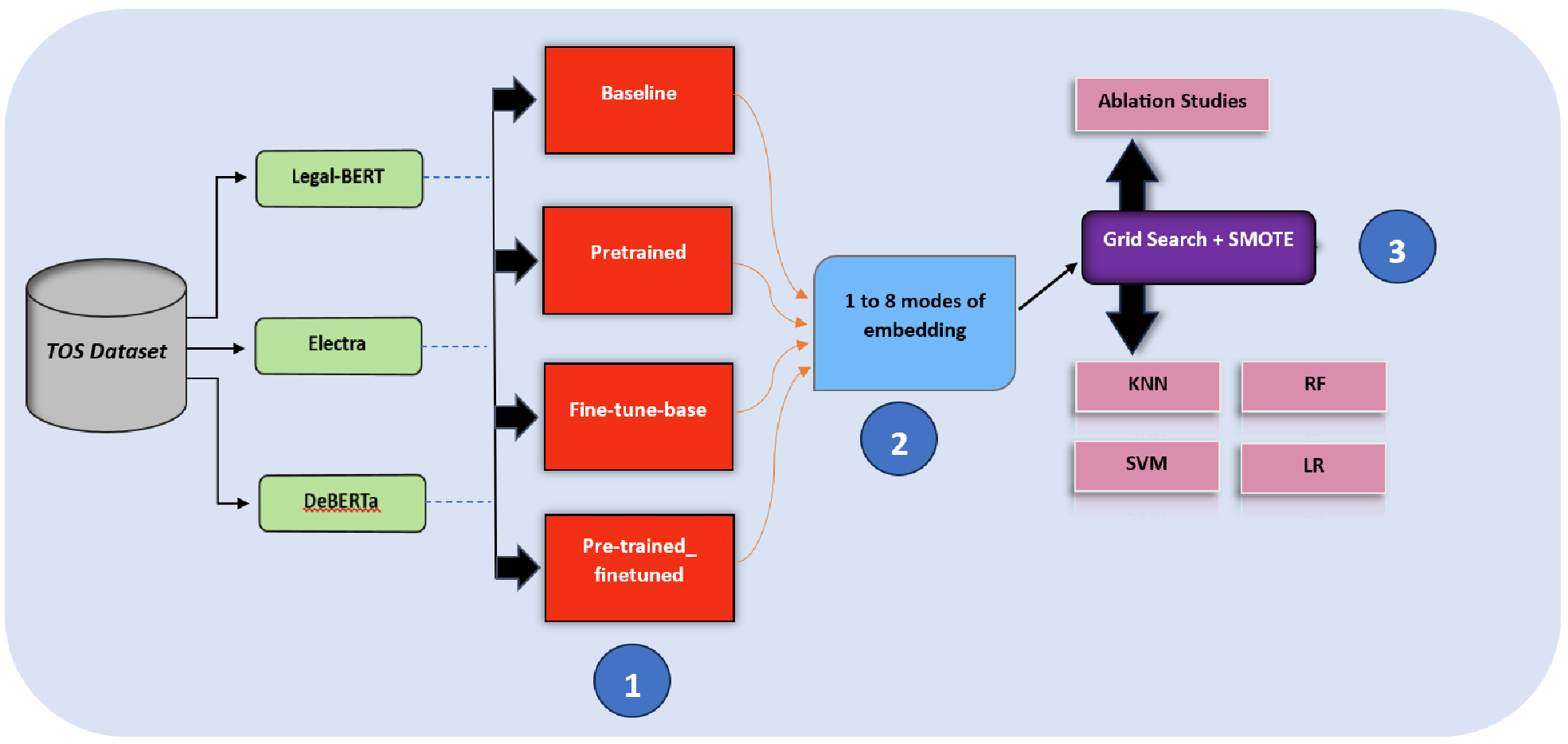}

\caption{Step by step Research Framework}
\label{fig:res_pipe_full}
\end{figure}

\begin{figure}[!h]
\centering
\includegraphics[scale=0.68]{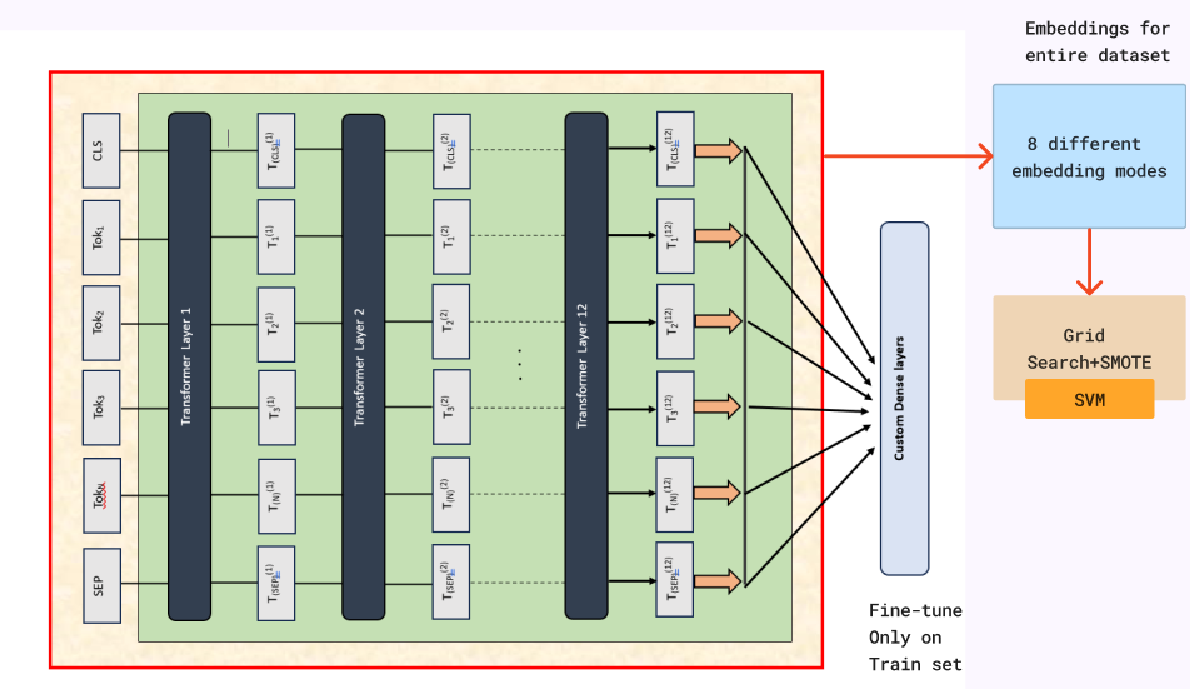}

\caption{Fine-tune-base Methodology}
\label{fig:res_pipe_full}
\end{figure}

\subsection{\textbf{Research Pipeline}}
% \vspace{-5pt}
The research pipeline is shown in Figure(\ref{fig:res_pipe_full}) The steps followed in the pipeline are as follows:

\begin{itemize}
    \item \textbf{Step-1}\textit{(Implementation of Methodology}): First, the dataset was split into testing and training with 0.15 split for testing. The test and train sets remain separated henceforth. This is because \textit{Fine-tune-base} approach utilizes only a training set. Any dilution of data could lead to inflation in results.  The first three aforementioned methodologies(Baseline, Pretrained, Fine-tune-base) were formulated for each transformer model, leading to nine combinations of output models (Legal-BERT*3 methodologies, Electra*3 methodologies, and DeBERTa*3 methodologies).  Added to this, Only Legal-BERT was processed with \textit{Pretrained\_finetuned} owing to the reasoning presented in Section(\ref{legal_bert_expla}). This led to the generation of 10 custom transformer models in total.  

    \item \textbf{Step-2}\textit{(Embedding generation)}: Next, sentence embeddings were calculated for each TOS dataset entry. While sentence embeddings are calculated from word embeddings, all eight modes were utilized. Hence, each transformer approach variant resulted in 8 different outputs. In total, 80 embedding sets were generated(10 Transformer models times eight embedding modes).

    \item \textbf{Step-3}\textit{(Grid search+SMOTE)}: There is a severe class imbalance in the TOS dataset in all categories, as illustrated in Table(\ref{data_desc}). In order to compensate, we utilize vanilla SMOTE (Synthetic Minority Over-sampling Technique)\cite{chawla2002smote} only on the training set. Grid Search was performed for hyper-parameter tuning of ML classifiers over all embedding sets generated in a 5-fold cross-validation fashion respectively. SMOTE was applied in the grid search pipeline for the training set alone. Finally, during testing the best fit hyperparameters were trained on training data after the application of SMOTE and tested on a pristine test dataset. 
 
\end{itemize}

% \begin{center}
%     RQ1: \emph {How does the performance of Transformer models fluctuate based on the approach utilized?} \\
%     RQ2: \emph{What is the extent of performance variation among distinct embedding modes  concerning Transformer models?} \\
%     RQ2: \emph{In relation to different embedding modes, what is the variability in performance observed across different Approches?} \\
% \end{center}
% \renewcommand{\labelitemi}{$\circ$}

% \begin{list}{$\circ$}{}  
% \item \textbf{Step-1}:  
% \item \textbf{Step-2}
% \end{list}

%% file: Sections/results_desc.tex
% \vspace{-10pt}
\vspace{-15pt}
\section{\MakeUppercase{Experiments and Results}}
\vspace{-2pt}
The study works on two tasks. The first task is a classification of each statement into unfair or not based on Unfair Binary as shown in Table(\ref{data_desc}). This problem checks for the existence of any clause in a statement but not for a specific unfair tag. The second task aims to categorize statements into a specific unfair tag. Each statement could have multiple unfair tags in different categories. Following the pipeline shown in Figure(\ref{fig:res_pipe_full}) steps 1 through 4 were followed. On initial testing with different ML classifiers on embeddings with baseline Legal-BERT embeddings using grid search, it was found that SVC clearly performed significantly better by a notable margin as compared to the other models. Hence, for all analysis and testing, only SVM(SVC) was used. Added to this, the performance of MLP was already tested in Section(\ref{custom_finetune}). 

SOTA results for unfair clause classification(Binary) are shown in Table(\ref{SOTA_unfair}). The metrics shown are macro testing averages for precision, recall, and F-score. Further, we also show other notable results from the pipeline for holistic discussion and comparison. A grid search was performed with 5-fold cross-validation in all cases, and SMOTE was applied only to the training data. For best fit, the kernel remained Radial Basis Function(rbf). The other hyperparameters are shown in Table(\ref{SOTA_unfair}). Further, the best-fit Mode for embedding is mentioned in '[]' where \textit{'SS'} refers to the use of SVC after Standard Scaler and \textit{'S'} refers to Vanilla SVC in the Grid Search pipeline respectively. The corresponding number is the Embedding Mode number. 

\begin{flushleft}
\begin{table*}[!h]
\caption{ Macro testing metrics on Unfair Binary }\label{SOTA_unfair}
\centering
\scalebox{0.76}{
\begin{tabular}{|@{\vrule width0pt height8pt\enspace}c|c | c | c |c | c |}\hline

\hline
\bf Transformer model & \bf Methodology   &\bf (c,gamma)[model] &\bf P &\bf R  &\bf F1 \\\hline
Legal-BERT &  Baseline & (10,0.001) [SS-2] & 0.891 & 0.836 & 0.86 \\
ELECTRA &  Baseline & (1,auto) [SS-7] & 0.853 & 0.837 & 0.845 \\
Legal-BERT &  Pretrained & (10,0.1) [S-1] & 0.872 & 0.875 & 0.874 \\
 
Legal-BERT  &  Pretrained-Fine-tuned & (10,0.001) [S-4] &  0.923 & \textbf{0.921} & \textbf{0.922} \\ 
ELECTRA &  Fine-tune-base & (1,0.001) [SS-8] &  \textbf{0.935} & 0.897 & 0.915 \\
DeBERTa &  Fine-tune-base & (100,0.01) [S-2] & 0.923 & 0.908 & 0.915 \\
\textbf{Legal-BERT } &  \textbf{Fine-tune-base} & (1,0.001) [SS-4] & 0.937 & 0.907 & \textbf{0.921} \\
% \hline
% \bf Work & \bf Approach   &  &\bf Precision &\bf Recall  &\bf F-Score 
\hline
\cite{lippi2019claudette} & Ensemble & & 0.828 & 0.798 & 0.806 \\

\cite{guarino2021machine} & SVM+mUSE & & 0.86 & 0.86 & 0.87  \\
\cite{zheng2021does} & Legal-BERT+Dense(softmax) & & - & - & 0.787 \\
\cite{mamakas2022processing} & LegalLongformer & & - & - & 0.806 
 
% Loos Et al.\cite{loos2016wanted} & SVM+mUSE & & 0.86 & 0.86 & 0.87 

\\\hline
\end{tabular}}
\end{table*}
\end{flushleft}

\vspace{-12pt}
\subsubsection{\textbf{Analysis and Discussion}}
\vspace{2pt}
The results surpass the current SOTA(State Of The Art), \cite{guarino2021machine} achieving macro F-1 at the binary unfair classification of 0.92. Even though Legal-BERT \textit{Pretrained-Fine-tuned} and \textit{Fine-tune-base} have comparable results, we propose Legal-BERT-Fine-tune-base with SVC for the problem(shown in bold above). The procedure is less elaborate and could be implemented with lesser computational capabilities. Pre-training consumes high GPU resources and has and performs worse asymptotically during the training phase. The final proposed design is illustrated in Figure(\ref{fine_tune_arch}). 

% \vspace{-4pt}
\subsection{\textbf{Experiments on each Unfair Category}}
% \vspace{-5pt}
The following section shows the results for the second task mentioned. However, in this case, we only consider the statements with at least one unfair clause (essentially unfair Binary annotated 1 as in Section(\ref{dataset_disc})). Further, each statement can have multiple unfair tags from different categories. This gives 1137 statements. On application of the design proposed above with Legal-BERT-Fine-tune-base in conjunction with SVC standard scaler and shown in \ref{fine_tune_arch}, the results obtained are shown below. All eight embedding modes are tested with Legal-BERT-Fine-tune-base, and the best results are shown. The study achieves SOTA results in all fields by a notable margin as compared to Lippi et al. and Guarino et al. Micro($\mu$) and macro(m) averages for F1-score are shown for comparison.

%  \begin{adjustwidth}{-6cm}{}
 
% \hskip-4.0cm \begin{table}[!h]
% \caption{Testing metrics based on Unfair clause category}\label{table10}
% \centering
% % \renewcommand{\arraystretch}{1.8}
% \begin{tabular}{|@{\vrule width0pt height8pt\enspace}c|c | c | c |c | c |c| }\hline

% \hline
%  Category   &\bf (c,gamma)[model] &\bf Precision &\bf Recall  &\bf F-Score(m,$\mu$) &\bf Lippi($\mu$F1) & \bf Guarino($\mu$F1)\\\hline 

% Arbitration & (1,0.01)[SS-1] & 1.0 & 1.0 & \textbf{1.0},1.0 & 0.823 & 0.89  \\ 
% Choice of law& (0.1,0.01)[SS-1] & 1.0 & 1.0 & \textbf{1.0},1.0 & 0.93 & 0.97 \\
% Contract by using & (0.1,0.01)[SS-1]& 0.989 & 0.929 & \textbf{0.956}.0.98 & \textbf{0.953} & 0.86 \\
% Content removal & (0.1,0.001)[SS-6] & 0.955 & 0.955 & \textbf{0.955},0.98 & 0.745 & 0.64  \\
% Jurisdiction & (0.1,0.01)[SS-1]& 1.0 & 1.0 & \textbf{1.0},0.99 & 0.970 & 0.97   \\
% Limitation of liability & (0.1,0.01)[SS-1] & 0.988 & 0.958 &\textbf{0.972},0.98 & 0.932 & 0.93\\ 
% Unilateral change & (0.1,0.01)[SS-1] & 0.94 & 0.894 & \textbf{0.914},0.94 & 0.823 & 0.86  \\
% Unilateral termination & (0.1,0.001)[SS-1] & 0.938 & 0.938 & \textbf{0.938},0.96 & 0.853 & 0.86 \\

% \hline
% \end{tabular}
% \end{table}
% \end{adjustwidth}

 \begin{table*}[!h]
\caption{Testing metrics based on each Unfair Clause/Category}\label{table10}
\centering
\scalebox{0.78}{
\begin{tabular}{|@{\vrule width0pt height8pt\enspace}c|c | c | c |c | c |c| }\hline

\hline
 Category   &\bf (c,gamma) &\bf P &\bf R  &\bf F1(m,$\mu$) &\bf Lippi($\mu$F1) & \bf Guarino($\mu$F1)\\\hline 

Arbitration & (1,0.01) & 1.0 & 1.0 & \textbf{1.0},1.0 & 0.823 & 0.89  \\ 
Choice of law& (0.1,0.01) & 1.0 & 1.0 & \textbf{1.0},1.0 & 0.93 & 0.97 \\
Contract by using & (0.1,0.01)& 0.989 & 0.929 & \textbf{0.956}.0.98 & 0.953 & 0.86 \\
Content removal & (0.1,0.001) & 0.955 & 0.955 & \textbf{0.955},0.98 & 0.745 & 0.64  \\
Jurisdiction & (0.1,0.01)& 1.0 & 1.0 & \textbf{1.0},0.99 & 0.970 & 0.97   \\
Limitation of liability & (0.1,0.01) & 0.988 & 0.958 &\textbf{0.972},0.98 & 0.932 & 0.93\\ 
Unilateral change & (0.1,0.01) & 0.94 & 0.894 & \textbf{0.914},0.94 & 0.823 & 0.86  \\
Unilateral termination & (0.1,0.001) & 0.938 & 0.938 & \textbf{0.938},0.96 & 0.853 & 0.86 \\

\hline
\end{tabular}}
\end{table*}

% \begin{table}
% \centering
% \caption{Accuracy Scores of Multi-Layer perceptron without SMOTE resampled dataset}
% \begin{tabular}{|c|c|c|c|c|c|}
% \hline
% Trait & Number of layers & Neurons per layer & Activation per layer & Learning rate & Test Accuracy \\
% \hline
% Extroversion & 3 & 200, 300, 300 & relu, relu, elu & 0.002 & 0.755 \\
% Openness & 3 & 100, 500, 400 & elu, relu, selu & 0.01 & 0.812 \\
% Agreeableness & 2 & 400, 100 & elu, selu & 0.003 & 0.729 \\
% Consciousness & 2 & 100, 200 & elu, elu & 0.007 & 0.651 \\
% Neuroticism & 2 & 100, 200 & relu, relu & 0.007 & 0.915 \\
% \hline
% \label{mlp_unsmote}
% \end{tabular}
% \end{table}

% \begin{table}[!h]
% \caption{ SOTA unfair clause detection results}\label{table10}
% \centering
% % \renewcommand{\arraystretch}{1.8}
% \begin{tabular}{|@{\vrule width0pt height8pt\enspace}c|c | c | c|}\hline

% \hline
% \bf ML classifier &\bf Precision &\bf Recall  &\bf F-Score \\\hline
% KNN  & 0.891 & 0.836 & 0.86 \\
% LR  & 0.853 & 0.837 & 0.845 \\
% SVM & 0.891 & 0.836 & 0.86 \\ 
% RF   &  0.7 & 0.921 & 0.922 \\ 

% \hline
% \end{tabular}
% \end{table}

%% file: Sections/case_study.tex
\vspace{-10pt}
\section{\MakeUppercase{Comparitive Analysis}}
\vspace{-3pt}
As shown in the Research Framework\ref{resframework}, numerous models and methodologies were utilized. In this Section we perform comparative analysis to answer two research questions. Although the results pertain to TOS data, they could be generically applied to Binary text classification problems and further help the research community in perspective and design. Based on the results generated at Unfair Binary Classification, the following research questions are answered:

\begin{center}
    \textbf{RQ1}: \emph {How does the performance of Transformer models fluctuate based on the methodology utilized?} \\
    \textbf{RQ2}: \emph{What is the extent of performance deviation among distinct embedding modes concerning Transformer models? How does the same vary depending on the methodology employed?} \\
\end{center}

The Research questions are answered in the sections below with empirical evidence based on the study.
% \vspace{-4pt}
\subsection{\textbf{RQ1}}
\vspace{4pt}

In order to reach a conclusion, the influence of each research methodology is analyzed in relation to each respective transformer model.

% \textbf{RQ1: \emph{How does the performance of Transformer models fluctuate based on the approach utilized?}}
\label{legal_bert_expla}
\textbf{Analysis based on Descriptive Statistics}: After the pipeline resulted in outputs for every combination, we first cluster the results based on each respective transformer model. Next, we provide the mean and max macro F-1 score as shown in table\ref{rq1}. In the case of each of these, the influence of embedding mode was deliberately omitted for a more focused understanding of the ablation regarding RQ1. Its to be noted that, Pretrained-Finetuned methodology was only calculated for Legal-BERT, this was due to the poor performance of other pre-trained transformer models as shown. The Descriptive Statistics lead to the following observations:
\vspace{-6pt}
\begin{enumerate}
  \item Baseline Legal-BERT appears to be the most versatile baseline model in light of its performance in varying conditions. 
  \item The performance of all models undergoing Fine-tune-base is very similar. Further, Legal-BERT performs slightly better at max statistics.
  \item In the case of pretraining, Legal-BERT clearly outperforms other models with Electra showing deteriorated results in mean and max statistics. DeBERTa and ELECTRA perform similarly at pretraining in Max statistics. Nevertheless, the mean performance of Electra, attributed to variations in embedding modes, shows a lower value compared to DeBERTa.
\end{enumerate}
\vspace{-6pt}

\begin{table*}[!h]
\caption{ Mean and max macro F1 at Unfair Binary Classification(model w.r.t methodology)}\label{rq1}
\centering
\scalebox{0.88}{
\begin{tabular}{|@{\vrule width0pt height8pt\enspace}c|c | c | c |c | c | c |}\hline

\hline
\bf Transformer & \bf Baseline   & \bf Fin-tune-bse & \bf Pre-tran & \bf Pre-fin-tune & \bf Av. & \bf Stat.\\\hline

Legal-BERT & 0.847 & 0.92 & 0.863 & 0.917 & 0.887 & mean \\
Electra & 0.827 & 0.912 & 0.777 & - & 0.839 & mean  \\
deBERTa & 0.82 & 0.912 & 0.812 & - & 0.848 & mean  \\
Legal-BERT & 0.86 & 0.921 & 0.874 & 0.922 & 0.894 & max  \\
Electra & 0.845 & 0.915 & 0.817 & - & 0.859 & max  \\
deBERTa & 0.825 & 0.915 & 0.818 & - & 0.853 & max  \\
% Average & 0.82 & 0.912 & 0.812 & - & 0.848 &
\hline
\end{tabular}}
\end{table*}

% \begin{table*}[!h]
% \caption{  Max macro F1 at Unfair Binary Classification(model w.r.t methodology)}\label{maxrq1}
% \centering
% % \renewcommand{\arraystretch}{1.8}
% \begin{tabular}{|@{\vrule width0pt height8pt\enspace}c|c | c | c |c | c |}\hline

% \hline
% \bf Transformer model & \bf Baseline   & \bf Fine-tune-base & \bf Pretrained & \bf Pre-Fine-tuned & \bf Average \\\hline

% Legal-BERT & 0.86 & 0.921 & 0.874 & 0.922 & 0.894 \\
% Electra & 0.845 & 0.915 & 0.817 & - & 0.859 \\
% deBERTa & 0.825 & 0.915 & 0.818 & - & 0.853 \\
% % Average & 0.82 & 0.912 & 0.812 & - & 0.848 &
% \hline
% \end{tabular}
% \end{table*}

Through this, we infer that although the performance of the transformer models in the case of Fine-tune-base is comparable, the study suggests that there is a notable difference in Pretraining and vanilla baseline performances. The best performance is shown in Fine-tune-base.

% \vspace{-4pt}
\subsection{\textbf{RQ2}}
\vspace{4pt}
The impact of each embedding method in relation to each individual transformer model and research methodlogy is examined independently in this section.

% \textbf{RQ1: \emph{How does the performance of Transformer models fluctuate based on the approach utilized?}}
\label{legal_bert_expla}
\textbf{Analysis based on Descriptive Statistics}: To answer this research question we first cluster the results based on each respective Embedding mode. We further show two variants in taking mean statistics, one takes mean w.r.t transformer models and the latter takes mean w.r.t research methodology.  The descriptive statistics over mean macro F1-score are shown in table\ref{rq1} leading to the following observations:

\begin{table*}[!h]
\caption{ Mean macro F1 at Unfair Binary Classification(Embedding Modes)}\label{table10}
\centering
\scalebox{0.85}{
\begin{tabular}{|@{\vrule width0pt height8pt\enspace}c|c |c|c|c|c|c|}\hline

\hline
\bf Mode & \bf Legal-BERT   & \bf Electra & \bf DeBERTa & \bf Baseline   & \bf Fin-tune-bse & \bf Pre-tran   \\\hline

1 & 0.889 & 0.822 & 0.844 &  0.832 & 0.914 & 0.8  \\
2 & 0.888 & \textbf{0.85} & 0.846 & 0.83 & 0.913 & \textbf{0.832} \\ 
3 & 0.884 & 0.844 & \textbf{0.852} & 0.832 & 0.915 & 0.821 \\
4 & 0.885 & 0.838 & \textbf{0.85} & 0.831 & 0.915 & 0.816 \\
5 & 0.889 & 0.82 & 0.844 & 0.829 & 0.914 & 0.8 \\
6 & 0.888 & \textbf{0.85} & 0.845 & 0.83 & 0.912 & \textbf{0.832} \\ 
7 & 0.884 & 0.845 & \textbf{0.852} & 0.834 & 0.915 & 0.821 \\
8 & 0.887 & 0.84 & 0.85 & 0.832 & 0.916 & 0.818 \\
\hline
\end{tabular}}
\end{table*}
\vspace{-6pt}
\begin{enumerate}
  \item There is little deviation in Legal-BERT performance in relation to embedding modes. Similarly, comparable deviation is shown in DeBERTa performance.
  \item ELECTRA shows maximum fluctuation with the best performance shown in embed modes 2 and 6 respectively. Embedding modes 1 and 5 perform notably lower at 0.82 macro F1.  
  \item For the results combining all Baseline Models and Fine-tune-base models, there is minimal variation.
  \item Finally, the mean Pretrained Model performance differs notably with Modes 2 and 6 performing the best. Modes 1,4 and 5 underperform relatively.
  
\end{enumerate}
Both modes 2 and 6 utilize mean embeddings of all layers and show better performance in the case of pre-training. The reason for this could be that holistic semantic information from all layers could be essential for text classification. The research by \cite{jawahar2019does} demonstrates that a complex hierarchy of linguistic information is encoded in BERT's intermediate layers, with semantic attributes at the top, surface characteristics at the bottom, and syntactic features in the middle layers.\cite{kovaleva2019revealing} validated a similar view, proving that the intermediary layers of BERT capture meaningful linguistic information.  \cite{zhong2023revisiting}, expounds on Semantic drift and Semantic loss that appears with BERT pre-training. Through this, we deduce that in the case of pretraining transformer-based models, embedding modes 2 and 6 perform relatively better. Both modes utilize a mean of all layers. In the case of Fine-tune-base and Baseline models, the performance seems to be similar.

% Analogously, \cite{kakouros2023does} checked for inclusion of Prosody in BERT structural information, showing its spread over several levels with large concentrated in the middle layers of the BERT. This was attributed to showing a dependence on syntactic and semantic signals.

% For each of these, influence of embedding mode was intentionally excluded for ablation understanding of the question.

% \textbf{Thus we deduce that, there is no notable difference in Openess, Concsceinceness and Extroversion based on zodiac signs altogh there is slight variation. In the case of agreeableness 3 mentioned zodiac pairs have significant difference with one pair for Neuroticism }. 

%% file: Sections/conclusion.tex
\vspace{-7pt}
\section{\MakeUppercase{Conclusion}}
\vspace{-2pt}
End-users predominantly fail to comprehend the complete consequences of terms they are consenting to owing to complexities in the language and a paucity of legal expertise. As a result, they frequently forgo reviewing these components. This enables firms to use ambiguous, complex, and unjust contractual provisions that restrict their own obligations or grant them the ability to capriciously terminate services they provide at an instance. Considering these complexities, we propose a Legal-BERT plus SVC-based approach to achieving proficient results at automated unfair clause detection. Comparative analysis is also performed on the various transformer architectures used. As future work to the study, we would aim to utilize other LLM approaches and prompt engineering to analyze contractual terms in the B2B market. These transactions between businesses and their contractual relations have been under light but with no significant results.

%% file: Sections/appendix.tex
\vspace{-10pt}
 \section{\MakeUppercase{Appendix}} \label{appendix}
% \vspace{-5pt}
%  % \thispagestyle{empty}

% Descriptions of the Dataset utilized are shown below.
% \vspace{-5pt}
\subsection{Availability of Code}
\vspace{-5pt}
The code and output performance metrics utilized in the study are available at  \href{https://github.com/batking24/Unfair-TOS-An-Automated-Approach-based-on-Fine-tuning-BERT-in-conjunction-with-ML}{Github} on request.